\documentclass[10pt,twocolumn,letterpaper]{article}

\usepackage{cvpr}
\usepackage{times}
\usepackage{epsfig}
\usepackage{graphicx}
\usepackage{amsmath}
\usepackage{amssymb}
\usepackage{dblfloatfix}
\usepackage{multirow}
\usepackage[pagebackref=true,breaklinks=true,letterpaper=true,colorlinks,bookmarks=false]{hyperref}

\cvprfinalcopy % *** Uncomment this line for the final submission

\ifcvprfinal\fi
\begin{document}

%%%%%%%%% TITLE
% \title{UniPose: Unified Semantic Pose Machine for Temporal Human Pose Estimation}
\title{
% UniPose: Unified Human Pose with Temporal Estimation
UniPose: Unified Human Pose Estimation in Single Images and Videos
}

\author{Bruno Artacho \hspace{2cm} Andreas Savakis \\
Rochester Institute of Technology\\
Rochester, NY\\
{\tt\small bmartacho@mail.rit.edu \hspace{0.5cm} andreas.savakis@rit.edu}}

\maketitle

\begin{abstract}
We propose UniPose, a unified framework for human pose estimation, based on our ``Waterfall" Atrous Spatial Pooling architecture, that achieves state-of-art-results on several pose estimation metrics.
Current pose estimation methods utilizing standard CNN architectures heavily rely on statistical postprocessing or predefined anchor poses for joint localization. 
UniPose incorporates contextual segmentation 
% of the entire image 
and joint localization to estimate the human pose in a single stage, with high accuracy, without relying on statistical postprocessing methods. 
The Waterfall module in UniPose leverages the efficiency of progressive filtering in the cascade architecture, while maintaining multi-scale fields-of-view comparable to spatial pyramid configurations.
Additionally, our method is extended to UniPose-LSTM for multi-frame processing and achieves state-of-the-art results for temporal pose estimation in Video. 
% The linear and 
% The sequential UniPose, with Long-Short Term Memory configuration for the Waterfall architecture, demonstrates high accuracy for temporal joint detection without significantly increasing the size of the network.
Our results on multiple datasets demonstrate that  UniPose,  with a ResNet backbone and Waterfall module, is a robust and efficient architecture for pose estimation obtaining state-of-the-art results in single person pose detection for both single images and videos.
% We also introduce a large high resolution dataset for pose estimation in American Sign Language, which consists of a variety of individuals signing in several different backgrounds. 
% This new dataset allows the tailored training and testing of pose estimation of the network for the specific task of Sign Language detection, allowing a more precise and insightful study of sign language.
\end{abstract}

\section{Introduction}
Human pose estimation is an important task in computer vision with applications in activity recognition \cite{ActionRecognition}, human computer interaction \cite{Gesture4HCI}, animation \cite{Avatar}, gaming \cite{Kinect}, health \cite{RehabPose}, and sports \cite{SportPerformance}. 
% However, pose estimation is challenging due to the high complexity caused by the number of degrees of freedom in the human body mechanics and the frequent occurrence of parts occlusion.
% Although, the pose estimation task receives ample investment and effort since its wide range of possible of applications and tasks such as activity recognition, animation, gaming, and recognition.
% The significance of pose estimation, in both the development of novel architectures and its practical use, 
The importance of pose estimation
has motivated the development of several approaches, in 2D \cite{DeepPose}, \cite{EfficientPoseCNN}, \cite{HourGlass}, \cite{HRNet} and 3D \cite{LCR-Net}, \cite{Monocap}, \cite{DensePose}; on a single frame \cite{RecurrentPose} or a video sequence \cite{3D_Video_Occlusion-Aware}; for a single \cite{CPM} or multiple subjects \cite{OpenPose}.

Pose estimation is challenging due to
% high complexity caused by the
the large number of degrees of freedom in the human body mechanics and the frequent occurrence of parts occlusion.
% Most architectures currently utilized for pose estimation 
% don't focus on the environment and contextual information of the frame in order to estimate the human pose, 
To overcome problems with occlusion, many methods rely on statistical and geometric models to estimate occluded joints \cite{GeometricPose}, \cite{StatisticalPose}. Another approach is the utilization of a library of known poses, known as anchor poses \cite{LCR-Net}, but this could limit the generalization power of the model and the ability to learn for unforeseen poses.

\begin{figure}[t]
\begin{center}
\includegraphics[width=1\linewidth]{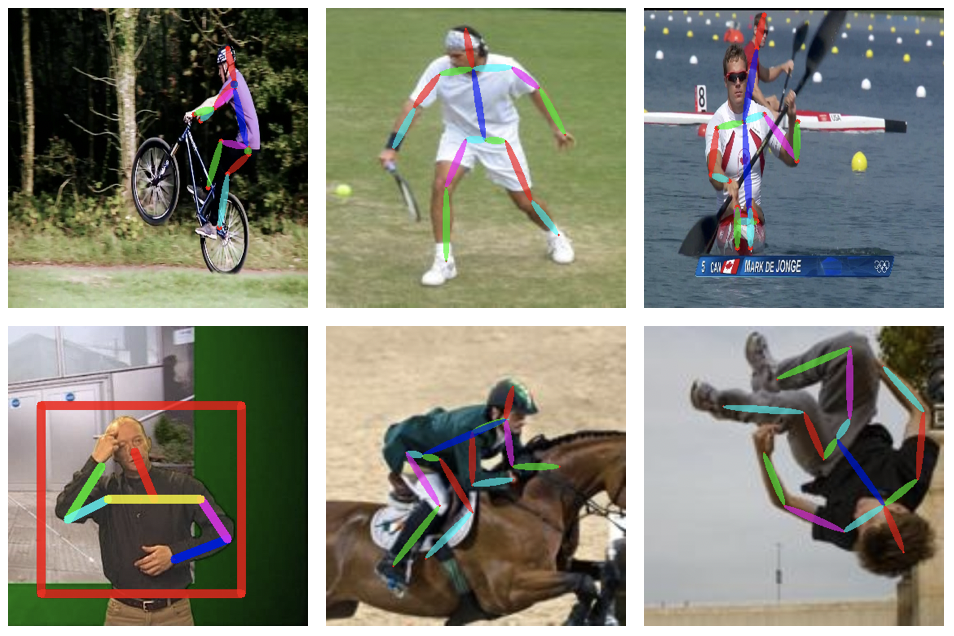}
\end{center}
  \caption{Pose estimation examples with our UniPose method.}
\label{fig:samples}
\end{figure}

Motivated by advances in semantic segmentation architectures \cite{Rethinking}, \cite{DilatedConv}, \cite{Enet}, 
% and early applications of semantic models for pose \cite{SemanticPoseMachine} 
% we propose the use of Atrous convolutions applied in a spatial pooling configuration for 
we propose a unified pose estimation framework, called UniPose, that consists of only one stage and obtains accurate results without postprocessing.
A main component of our architecture is the Waterfall Atrous Spatial Pooling (WASP) module
% , shown in Figure \ref{fig:WASP}, 
which combines the cascaded approach for Atrous Convolution with the larger FOV obtained from parallel configuration from the Atrous Spatial Pyramid Pooling (ASPP) module \cite{DeepLab}. 

Our unified approach predicts the location of joints using contextual information due to the larger Field-of-View (FOV) and multi-scale approach used in our network. 
With our contextual approach, our network includes the information of the entire frame and, therefore, does not require post analysis based on statistical or geometric methods. Examples of pose estimation obtained with our UniPose method are shown in Figure \ref{fig:samples}.
The main contributions of this paper are the following.
\begin{itemize}
\vspace{-0.07in}
\item We propose the UniPose framework, based on the Waterfall module for Atrous Spatial Pooling, that achieves state-of-the-art results for single person human pose estimation.
\vspace{-0.08in}
\item Our Waterfall module increases the receptive field of the network by combining the benefits of cascade atrous convolutions with multiple FOV in a parallel architecture inspired by the spatial pyramid approach.
\vspace{-0.08in}
\item The proposed UniPose method determines both the  locations of joints and the bounding box for person detection, eliminating the need for 
separate branches in the network.
% an extra Region Proposal Network (RPN) in a separate branch of the network.
\vspace{-0.08in}
\item We extend the Waterfall based approach to UniPose-LSTM  by adopting a linear sequential LSTM configuration and obtain state-of-the-art results for temporal human pose estimation in video.
% \vspace{-0.08in}
% \item We introduce the PoseASL dataset consisting of 786,384 high resolution labelled images of ASL from several different signers in different environments. Our new dataset allows the improvement of pose estimation for signers, enabling further studies and analysis of sign languages.
\end{itemize}

\begin{figure*}[t]
\begin{center}
\includegraphics[width=1\linewidth]{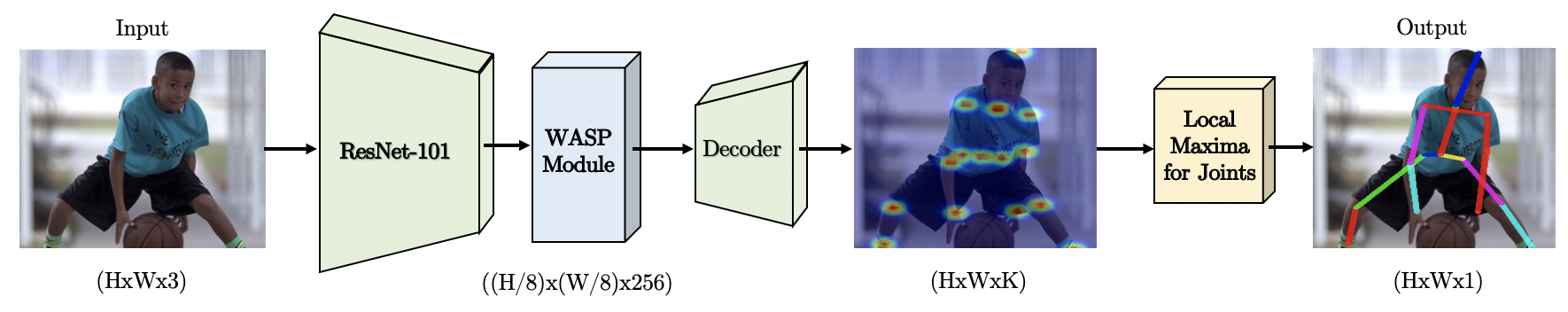}
\end{center}
  \caption{UniPose architecture for single frame pose detection. The input color image of dimensions (HxW) is fed through the ResNet backbone and WASP module to obtain 256 feature channels at reduced resolution by a factor of 8. The decoder module generates K heatmaps, one per joint, at the original resolution, and the locations of the joints are determined by a local max operation.}
\label{fig:unipose}
\end{figure*}

\section{Related Work}
Traditional methods for human pose estimation focused on the detection of joints, and consequently pose, via techniques that explored the geometry between joints in the target image \cite{Poselet}, \cite{Articulated}, and \cite{MixtureModels}.
In recent years, methods relying on Convolutional Neural Networks (CNNs) achieved superior results \cite{HRNet}, \cite{OpenPose}, \cite{LCR-Net}. 
% % As the main contribution of CNN for pose estimation, 
% The CNN's ability to generalize and learn different poses and spatial relations from data gives a decisive advantage over traditional geometric methods.
% Thus, deep CNN base models became the main methodology explored for modern pose estimation techniques:
The popular Convolutional Pose Machine (CPM) \cite{CPM} took an approach that refined joint detection via a set of stages in the network.
% This approach was used by the popular Convolutional Pose Machine (CPM) \cite{CPM}.
% , achieving great improvement to other methods available. 
Stacked hourglass networks \cite{HourGlass} utilized cascades of the hourglass structure for the pose estimation task.
Building upon \cite{CPM}, Yan et al. integrated 
% the iterative approach with 
the concept of Part Affinity Fields (PAF), resulting in the OpenPose method \cite{OpenPose}. 
PAF uses the detection of more significant joints to better estimate the prediction of less significant joints. This innovation allowed advances toward multi-person detection with decreased complexity and computational power.
% required in the detection and estimation.

% Introduced by Chu et al. \cite{Multi-context}
The multi-context approach in \cite{Multi-context} relies on an hourglass backbone to perform pose estimation. 
The original backbone
% receives the addition of 
is augmented by
the Hourglass Residual Units (HRU) with the goal of increasing the receptive FOV. 
% The method relies on 
Post processing with Conditional Random Fields (CRF) is used to assemble the relations between detected joints. However, the drawback of  CRF is increased complexity that requires high computational power and results in a reduction in speed.

% Differently than \cite{HourGlass} and \cite{Multi-context}, 
The High-Resolution Network (HRNet) \cite{HRNet} includes both high and low resolution representations. Starting with high resolution, the method gradually adds low resolution sub-networks to form more stages, and performs multi-scale fusion between sub-networks. 
% The approach in \cite{HRNet} 
HRNet benefits from the larger FOV of multi resolution, a capability that we achieve in a simpler fashion with our WASP module.

DeepPose \cite{DeepPose} utilizes a cascade of deep CNNs and locates body joints via regression. The method relies on iterative refinement in order to better predict symmetric and lower confidence joints.
% in a more holistic approach in a cascade fashion in order to discover better relations between joints in the image.

% More recently, work has been conducted aiming to
Some recent works attempt to leverage contextual information into pose estimation. The Cascade Prediction Fusion (CPF) \cite{zhang2019human} uses graphical components in order to exploit the context for pose estimation. Similarly, the Cascade Feature Aggregation (CFA) \cite{su2019improvement} aims to use semantic information to detect pose with a cascade approach.

The Location, Classification, and Regression network (LCR-Net) \cite{LCR-Net} extends pose estimation to 3D space via depth regression. 
LCR-Net relies on a Detectron backbone \cite{Detectron} for the detection of human joint locations. 
From these locations, the method finds the best fit to predefined anchor poses for the detected human poses. Finally, LCR-Net performs a regression to estimate 3D coordinates in the image. A drawback of this method is the limited set of anchor poses available, which impose a limitation on the network for the estimation of unforeseen poses.

In a different approach for 3D pose estimation, the MonoCap method for human capture \cite{Monocap} couples a CNN with a geometric prior in order to statistically determine the third dimension for the pose using the Expectation-Maximization algorithm.

A drawback of some current methods is that they require an independent branch for the detection of the bounding box of human subjects in the frame. LightTrack \cite{LightTrack}, for instance, relies on a separate YOLO \cite{YOLOv3} architecture to perform the detection of subjects prior to detecting joints. In a different feamework, LCR-Net \cite{LCR-Net} has different branches for the detection using Detectron \cite{Detectron} and the arrangement of joints during classification.

\subsection{Temporal Pose Estimation}
For the task of pose estimation in videos, most methods 
% encounter difficulties and do not achieve a comparable result since they 
do not account for the temporal component and process each frame independently.
An additional challenge is
% can be affected by 
the occasional blurring resulting from the movement of the humans in the video. 
The main incentive for developing a pose estimation method that takes into account the temporal component is 
% that the method potentially will be able 
to better estimate joints during blurring or occlusion
% that are occluded in subsequent frames, 
using information from previous frames.

Targeting video applications, Modeep \cite{Modeep} utilized color channels from adjacent frames as input attempting to merge the motion in the video. Pfister et al. \cite{GesturePose} also proposed a similar technique to detect gestures in a video sequence.

More recently, optical flow techniques were adopted to tackle the temporal component for pose estimation. 
Deepflow \cite{Deepflow} used optical flow to better connect predictions between frames in a more continuous detection. Another method that utilized
% achieved expressive results with 
optical flow is Thin-Slicing \cite{Thin-Slicing}, relying on both optical flow and spatial-temporal model. 
However, the increased complexity of this model results in a significant increase in computational cost.

The Chained Model \cite{Chained} utilizes recurrent architectures to incorporate the temporal component. 
% The same concept from the Recurrent Neural Network (RNN) was also applied 
A similar concept was adopted 
by the LSTM Pose Machine \cite{LSTM-PM} approach, where the LSTM was utilized as the memory augmentation of the network.

Applications of LSTM aren't limited to the temporal component. 
Recurrent 3D Pose Sequence Machines (RSPM) \cite{RSPM} used LSTMs in the regression from 2D to 3D, 
to obtain better correspondence during the regression.

\subsection{Pose Estimation for Sign Language}
Despite the efforts on generic pose estimation methods, specific applications, such as for sign language, are currently lacking in research.
Charles et al. \cite{AutomaticSign} estimated pose during signing in long television broadcasting videos. The method relied on an initial separation from the background by the use of semantic segmentation, followed by a random forest regression to locate the upper limbs of the signer.
The work in \cite{UpperBodySign} used temporal tracking in order to detect parts and estimate upper body joints in similar frames. 
% The method also relies on sampling of pictorial structure proposal distribution for the determination of the output. 

DeepSign \cite{DeepSign} applied transfer learning on a pretrained CNN  for joint detection during sign language. Their approach followed the work done by \cite{DeepPose} in generic pose images and incorporated application specific transfer learning in the final architecture.

\subsection{Atrous Convolution and ASPP}
An important challenge with both semantic segmentation and pose estimation methods incorporating CNN layers is the significant reduction of resolution caused by pooling.
Fully Convolutional Networks (FCN) \cite{FCN} \cite{FCN} addressed the resolution reduction problem by deploying upsampling strategies across deconvolution layers.  
These attempt to reverse the convolution operation and increase the feature map size back to the dimensions of the original image.

A popular technique in semantic segmentation is the use of dilated or Atrous or dilated convolutions \cite{DeepLab}. The main objectives of Atrous convolutions are to increase the size of the receptive fields in the network, avoid downsampling, and generate a multi-scale framework for processing. 
% The name Atrous is derived from the French expression ``algorithm \`a trous", or translated to English ``Algorithm with holes". As alluded by its name, Atrous Convolutions alter the convolutional filters by the insertion of ``holes", or zero values in the filter, resulting in the increased size of the receptive field, resembling a hybrid of convolution and pooling layers.

% The use of Atrous Convolutions in the network is shown in Figure \ref{fig:Atrous}.

% \begin{figure}[htpb]
% \begin{center}
% \includegraphics[width=1\linewidth]{images/Atrous.png}
% \end{center}
%   \caption{Input pixels using a 3X3 Atrous Convolutios with different dilation rates of 1, 2, and 3, respectively.}
% \label{fig:Atrous}
% \end{figure}

In the simpler case of a one-dimensional convolution, the output of the signal is defined as follows:
\begin{equation}
    y[i]=\sum_{l=1}^{L}x[i+rl]\cdot w[l]
\end{equation}
\noindent 
where $r$ is the rate of dilation, $\omega[l]$ is the filter of length $L$, $x[i]$ is the input, and $y[i]$ is the output. A rate value of one results in a regular convolution operation.

Motivated by the success of the Spatial Pyramids applied on pooling operations \cite{SPP}, the ASPP architecture was successfully incorporated in DeepLab \cite{DeepLab} for semantic segmentation.
The ASPP approach assembles atrous convolutions in four parallel branches with different rates, that are combined by fast bilinear interpolation with an additional factor of eight. This configuration recovers the feature maps in the original image resolution.
% Previous applications of Atrous convolutions in semantic segmentation applications open possibilities of its use for other task, such as pose estimation. 
The increase in resolution and FOV in the ASPP network can be beneficial for a contextual detection of body parts during pose estimation. We leverage this capability in a more efficient manner with our Waterfall architecture in the UniPose framework.

\begin{figure*}[th]
\centering
\includegraphics[width=0.9\linewidth]{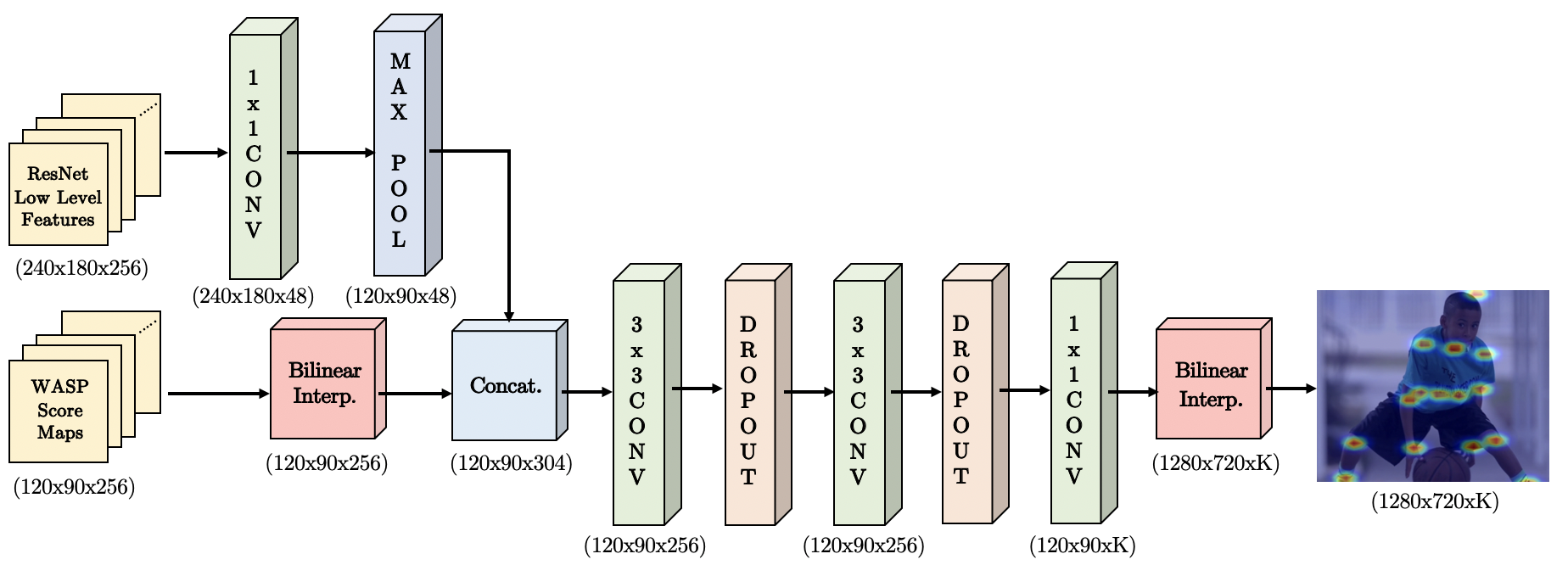}
\caption{Decoder module used in the UniPose pipeline. The original image dimensions are (1280x720). The inputs to the decoder are 256 channels of ResNet low level features and 256 channels of the WASP feature maps. The output of the decoder is K heatmaps corresponding to K joints, shown in the image example. Additionally, the decoder outputs heatmaps for the bounding box (not shown in the image). }
\label{fig:decoder}
\end{figure*}

\section{UniPose Architecture}
We propose UniPose, a unified architecture for pose estimation, that exploits the large FOV generated by atrous convolutions combined with cascade of convolutions in a ``Waterfall" configuration. 
Our WASP module 
% provides benefits due to its 
offers multi-scale representations as well as efficiency in the reduced size of the network. 
Improving upon previous works,  UniPose does not require separate branches for bounding box and joint detections. 
Instead, it performs a unified detection of the bounding box for the human subject and its joints.

The UniPose processing pipeline is shown in Figure \ref{fig:unipose}. The input image is initially fed into a deep CNN, in this case  ResNet-101, with the final layers replaced by a WASP module. 
The resultant feature maps are processed by a decoder network that generated K heatmaps, one for each joint, with the corresponding probability distributions obtained from Softmax. 
Then the decoder performs bilinear interpolation to recover the original resolution, followed by a local max operation to localize the joints for pose estimation.
The decoder in our network generates detections of joints for both visible and occluded parts. 
Additionally, the decoder generates a bounding box detection without the use of post-processing or independent parallel branches. 

% This unique and unified nature of our method qualifies it to the name of UniPose, an unified method for human pose estimation and detection.

We next provide the motivation for the development of the WASP module and contrast it with traditional deconvolutions in \cite{FCN} and the ASPP architecture in \cite{DeepLab}.

\begin{figure}[ht]
\begin{center}
\includegraphics[width=1\linewidth]{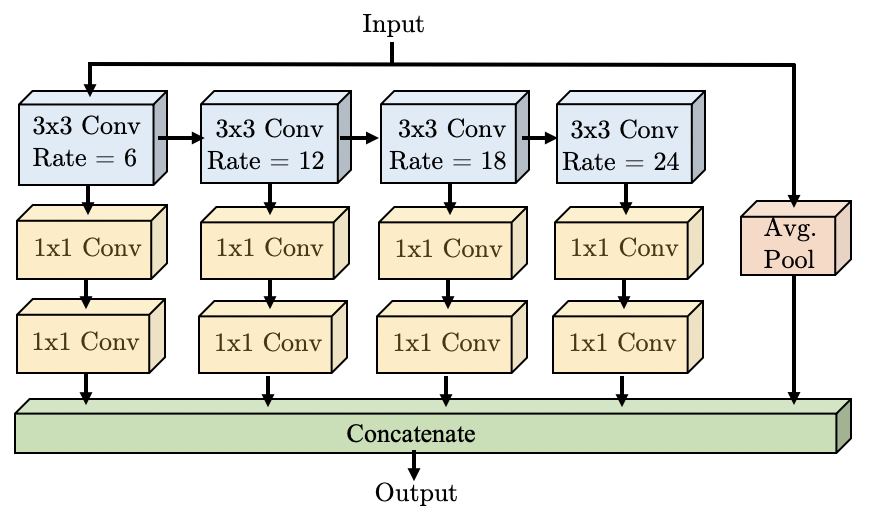}
\end{center}
  \caption{Waterfall architecture in the WASP module.}
\label{fig:WASP}
\end{figure}

\subsection{WASP Module}
The WASP module generates an efficient multi-scale representation that 
% is a novel architecture that 
helps UniPose achieve state-of-the-art results.
% in
% semantic segmentation and
% pose estimation. 
The WASP architecture, shown in Figure \ref{fig:WASP}, is designed to leverage both the larger FOV of the ASPP configuration and the reduced size of the cascade approach.
The inspiration for WASP was to combine the benefits of the ASPP  \cite{DeepLab}, Cascade  \cite{Rethinking}, and Res2Net  \cite{Res2Net} modules.

WASP relies on atrous convolutions, which are fundamental to ASPP, to maintain a large FOV. It also performs a cascade of atrous convolutions at increasing rates to gain efficiency, a concept motivated by the cascade approach. Furthermore, WASP incorporates multi-scale features inspired by the Res2Net architecture and other multi-scale approaches. In contrast to ASPP and Res2Net, WASP does not immediately parallelize the input stream. Instead, it creates a waterfall flow by first processing through a filter and then creating a new branch. WASP also goes beyond the cascade approach by combining the streams from all its branches and average pooling of the original input to achieve a multi-scale representation.

WASP is designed with the goal of reducing the number of parameters in order to deal with memory constraints and overcome the main limitation of atrous convolutions. 
The  four branches in WASP have different FOV and are arranged in a waterfall-like fashion.
The atrous convolutions in WASP start with a small rate of 6, which consistently increases in subsequent branches (rates of 6,12,18,24).  
This configuration gains efficiency due to the smaller filter sizes, and creates multi-scale features with each branch that are combined to obtain a richer representation.
The WASP module is utilized in the UniPose architecture of Figure \ref{fig:unipose} for pose estimation. 

\begin{figure*}[h]
\begin{center}
\includegraphics[width=1\linewidth]{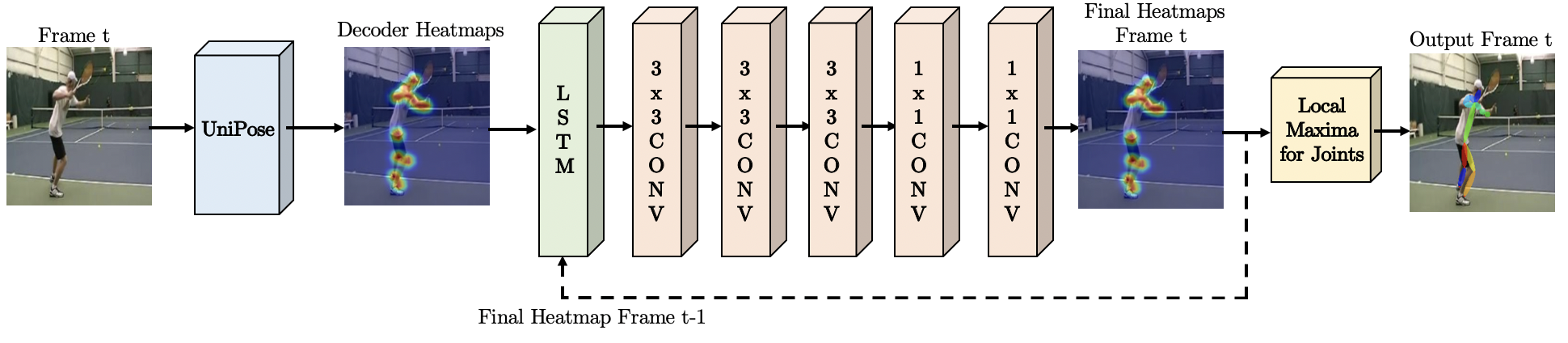}
\end{center}
  \caption{UniPose-LSTM architecture for pose estimation in videos.
  The joint heatmaps from the decoder of UniPose are fed into the LSTM along with the final heatmaps from the previous LSTM state. The convolutional layers following the LSTM reorganize the outputs into the final heatmaps used for joint localization. 
  }
\label{fig:unipose-LSTM}
\end{figure*}

\subsection{Decoder Module}
Our decoder module converts the score maps resulting from the WASP module to heatmaps corresponding to body joints and the bounding box. 
% The decoder receives WASP features and low level features. 
% The decoder is composed of convolutional layers, dropout layers, and bilinear interpolations to generate output maps at the same resolution as the input image.
Figure \ref{fig:decoder} shows the decoder architecture for an input color image of size (1280$\times$720).
% $\times$3 f
% or width, height, and RGB color channels, respectively. 
The decoder receives 256 feature maps from WASP and 256 low level feature maps from the first block of the ResNet backbone. After a max pooling operation to match the dimensions of the inputs, the feature maps are concatenated and processed through convolutional layers, dropout layers, and a final bilinear interpolation to resize to the original input size.
The output consists of K heatmaps corresponding to K joints that are used for joint localization after a local max operation. Additionally, the decoder outputs heatmaps for the bounding box without requiring an additional branch.

% \subsection{UniPose for Single Image}
% The single frame version of UniPose receives individual images and performs pose prediction. The method pipeline in Figure \ref{fig:unipose} shows an input image fed into a deep CNN backbone based on the ResNet-101 architecture, with the final layers replaced by a WASP module shown in Figure \ref{fig:WASP}. The resulting score map, as obtained from the Softmax probability distributions, is processed by a decoder, shown in Figure \ref{fig:decoder}.  The decoder performs bilinear interpolation and 2 layers of convolution for generating the score maps for all joints at the same resolution as the input image.

% The single image mode of UniPose also generates a bounding box containing the subject for pose estimation, in addition to the detection of the joints. This feature of the network eliminates the need for an RPN branch, that is used in other networks for the detection of the subject.

\subsection{UniPose-LSTM for Pose Estimation in Video}
The UniPose architecture was modified to UniPose-LSTM for pose estimation in video. 
For video processing, it is useful to leverage the similarities and temporal correlations between consecutive frames.
% , as well as the temporal correlations across these frames. 
% In order to better address this capability of our UniPose method, we present a modified version for the task of pose estimation in videos.
% To better address this enhancement of our UniPose method, we present a temporal version for the task of pose estimation in videos

To operate in video processing mode, the UniPose architecture is augmented by an LSTM module that receives the final heatmaps from the previous frame along with the decoder heatmaps from the current frame. 
The pipeline of UniPose-LSTM is shown in Figure \ref{fig:unipose-LSTM}.
This network includes CNN layers following the LSTM to generate the final heatmaps used for joint detection. 

The UniPose-LSTM configuration allows the network to use information from the previously processed frames, without significantly increasing the total size of the network. For both the single image and video configurations, our network uses identical ResNet-101 backbone, WASP module, and decoder.
We evaluated the performance benefits due to the temporal length of the memory component, when using an LSTM for several frames.
% , by considering different batch sizes for the network. 
It was experimentally determined that  accuracy improves when incorporating up to 5 frames in the LSTM, and a plateau in accuracy was observed for additional frames.
% Our results illustrate that performance gains were present when using up to 5 consecutive frames and a plateau in accuracy was observed for additional frames.

\section{Datasets}
We performed experiments on four datasets. Two of the datasets are composed of single images: Leeds Sports Pose (LSP) \cite{LSP} and MPII \cite{MPII}; and two datasets are composed of video sequences: Penn Action \cite{PennAction} and BBC Pose \cite{BBC}. A brief description of these datasets is provided below. 

The Leeds Sports Pose (LSP) dataset \cite{LSP} was initially used for single person pose estimation. Images for LSP were collected from Flickr for a variety of individuals performing sports activities. The dataset is composed of 1,000 images for training and 1,000 images for testing with 14 labelled keypoints in the entire body. The LSP dataset includes lower variation in the data, allowing a good initial assessment of the network performance for the task of single person pose estimation.

The MPII \cite{MPII} dataset contains approximately 25,000 images of annotated body joints of over 40,000 subjects. The images are collected from YouTube videos in 410 everyday human activities. The dataset contains frames with 2D and 3D joints annotations, head and torso orientations, and body part occlusions. Another feature of the MPII dataset is that it contains previous and following frames, although it lacks labelling for those frames.

Penn Action \cite{PennAction} dataset contains 2,326 video sequences of 15 different activities including different sports, athletic activities, and playing instruments. The dataset was used to evaluate the performance of our architecture for temporal pose estimation and joint tracking, i.e., the estimation of pose in a frame while contextually using previous detections to refine the result.

The BBC Pose dataset \cite{BBC} consists of 20 videos from the British Broadcasting Corporation (BBC) with the presence of a British Sign Language (BSL) signer. The BBC Pose dataset was utilized for the specialized application of human pose for sign language. 
The dataset includes of 610,115 labelled images for training, 309,171 for validation, and 309,260 for testing. As a limitation of the dataset, the labels consist of only 7 keypoints in the human upper body including head, shoulders, elbows, and wrists.

\subsection{Data Pre-Processing}
In order to train our network for joint detection, a pre-processing step was performed. 
Ideal Gaussian maps were generated at the locations of joints in the ground truth labels.
These maps are more effective for training than single points at the joint locations, and they are used to train our UniPose network to generate Gaussian heatmaps corresponding to the location of each joint in the frame.

Gaussians with different $\sigma$ values were considered in the training of the network to evaluate their effectiveness. For the presented results and final analysis, a value of $\sigma=3$ was adopted, resulting in a well defined Gaussian curve for both the ground truth and predicted outputs. This value of $\sigma$ also allows enough separation between joints in the image.

\section{Experiments}
We performed training, validation and testing of UniPose based on the procedures and metrics described in this section.  
Compared to state of the art, our methods achieved superior performance in several datasets, for both single frame pose estimation with UniPose and video pose estimation with UniPose-LSTM, including the specific task of pose estimation on sign language videos.

\subsection{Metrics}
For the evaluation of UniPose, various datasets and metrics were used, depending on previously reported results and the available ground truth for each dataset.
% The Average Precision metric (AP) is used as a baseline that measures the percentage of detections, for each joint, that are located at the same location as the ground truth.
Some datasets, such as LSP \cite{LSP}, report and compare accuracy in Percentage of Correct Parts (PCP), where a limb is considered detected if the distance of its two predicted joints is below a threshold. In this paper, we adopted a threshold of half the distance of the ground truth limb, commonly referred as PCP@0.5.
The PCP method introduces a bias due to the stronger penalization for the detection of smaller limbs (i.e. arm in comparison to torso), since they naturally have a shorter distance, and consequently a smaller threshold for detection.

Another metric used is the Percentage of Correct Keypoints (PCK). This metric considers the prediction of a keypoint correct when a joint detection lies within a certain threshold distance of the ground truth. 
Two commonly used thresholds were adopted. The first is PCK@0.2, which refers to a threshold of 20\% of the torso diameter, and the second is PCKh@0.5, which refers to a threshold of 50\% of the head diameter.

\subsection{Simulation Parameters}
We input the native resolution of the input image without resizing, in order to train the network with the most detail possible through our dense and large FOV network. 
For that reason, the batch size utilized varied from high amounts for lower resolution datasets (e.g. LSP) to smaller batches of 4 for datasets such as the BBC Pose \cite{BBC}.

We experimented with different rates of dilation on the WASP module. We found that larger rates result in better prediction. A set of dilation rates of $r =$ \{6, 12, 18, 24\} was selected for the WASP module.

We calculate the learning rate based on the step method, where the learning rate started at $10^{-4}$ and was reduced progressively by an order of magnitude at each step \cite{Parsenet}. 
All experiments were performed using PyTorch 1.0 running on Ubuntu 16.04. The workstation has an Intel i5-2650 2.20GHz CPU with 16 GB of RAM and an NVIDIA Tesla V100 GPU.

\section{Results}
% Following training, validation, and testing procedures, the UniPose architecture was implemented following the parameters and procedures described. 
% Initially, w
We initially tested our network on the LSP dataset and compared the results with other methods, as shown in Table \ref{tab:LSP}.
UniPose achieved a PCP of 72.8\% and a PCK@0.2 of 94.5\%,
% for detections with distance threshold of 20\% of the normalized torso diameter. 
% Our method shows 
showing significant gains in comparison to other approaches in both metrics.

% \begin{table}[!ht]
% \begin{center}
% \begin{tabular}{|c|c|c|}
% \hline
%     Architecture&PCP&PCK@0.2\\
% \hline\hline
%     \textbf{UniPose (ours)}&\textbf{72.8\%}&\textbf{94.5\%}\\
%     Part Regression \cite{PartHeatRegression}&-&90.7\%\\
%     CPM \cite{CPM}&-&90.5\%\\
%     DeepCut \cite{DeepCut}&-&87.1\%\\
%     Recurrent \cite{RecurrentPose}&-&85.2\%\\
%     DeepPose \cite{DeepPose}&61\%&-\\
%     Poselet \cite{Poselet}&56\%&-\\
%     Tian et al.\cite{MixtureModels}&56\%&-\\
% \hline
% \end{tabular}
% \end{center}
% \caption{LSP \cite{LSP} dataset average joints detection results}
% \label{tab:LSP}
% \end{table}

% Our results demonstrate a significant improvement from previous single person methods for 2D detection.
% Also, 
Differently than methods such as CPM \cite{CPM}, UniPose is able to detect the body joints with high confidence in a single iteration, instead of going through several stages or iterations in the network.

\begin{figure}[htbp]
\begin{center}
\includegraphics[width=1\linewidth]{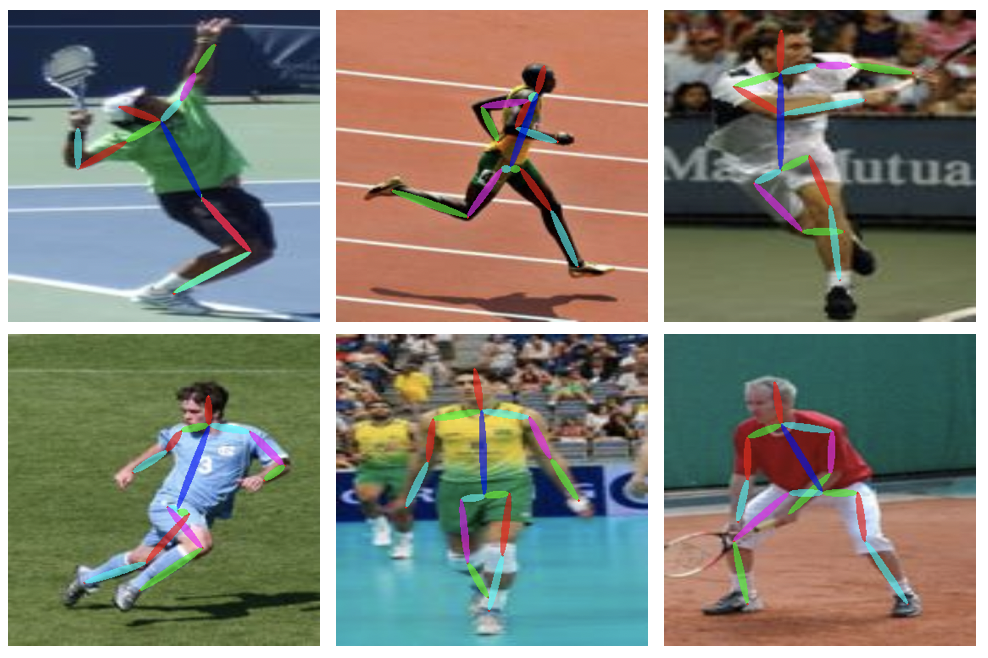}
\end{center}
  \caption{Pose estimation examples from the LSP dataset}
\label{fig:LSP_detection}
\end{figure}

% Overall, UniPose provides accurate joint detections that are superior to current state of the art methods. 
Examples of pose estimation for subjects from LSP dataset are shown in Figure \ref{fig:LSP_detection}.
% for testing images of the LSP dataset \cite{LSP}. 
It is noticeable from these examples that our method identifies the location of symmetric body joints with high precision. 
% A limitation that is observed in our detections, is a lower success rate to correctly assemble joints from not sufficiently separated limbs that also include occlusions.
Challenging conditions include the detection of joints in limbs that are not sufficiently separated and occlude each other.

\begin{table}[!ht]
\begin{center}
\begin{tabular}{|c|c|c|}
\hline
    \multirow{2}{*}{Method}&PCP&PCK@0.2\\
    &for LSP&for LSP\\
\hline\hline
    \textbf{UniPose (ours)}&\textbf{72.8\%}&\textbf{94.5\%}\\
    8-Stack HG \cite{zhang2019human}&-& 94.0\%\\
    Part Regression \cite{PartHeatRegression}&-&90.7\%\\
    CPM \cite{CPM}&-&90.5\%\\
    DeepCut \cite{DeepCut}&-&87.1\%\\
    Recurrent \cite{RecurrentPose}&-&85.2\%\\
    DeepPose \cite{DeepPose}&61\%&-\\
    Poselet \cite{Poselet}&56\%&-\\
    Tian et al.\cite{MixtureModels}&56\%&-\\
\hline
\end{tabular}
\end{center}
\caption{Pose estimation results and comparison with other methods 
% using Percentage of Correct Parts (PCP) and Percentage of Correct Keypoints (PCK@0.2) 
for the
LSP dataset.
}
% average joints detection results}
\label{tab:LSP}
\end{table}

% Following the initial tests, we 
We next perform training and testing in the larger MPII dataset \cite{MPII}, focusing on single person detection. Since the MPII images may contain multiple people, we used the center map of the main person in order to detect the pose of the correct individual. 
% To extract the information of a single individual, 
We used "Detectron2" \cite{Detectron} for segmentation and detection of all the individuals in the image, followed by the UniPose method to detect the pose of the selected individual.

\begin{figure}[htbp]
\begin{center}
\includegraphics[width=1\linewidth]{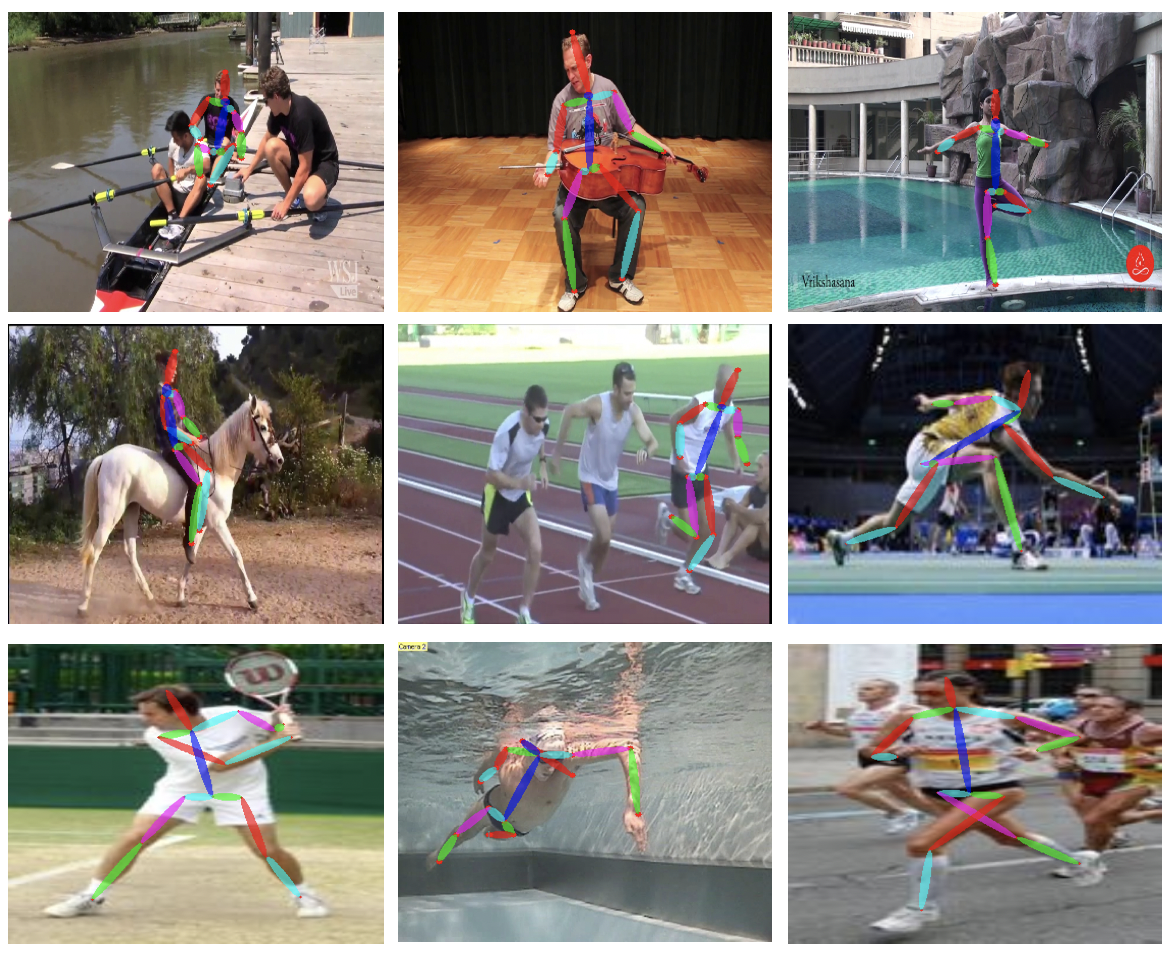}
\end{center}
  \caption{Pose estimation examples from the MPII dataset}
\label{fig:MPII_sample}
\end{figure}

Table \ref{tab:MPII} shows the results for the MPII testing dataset. UniPose achieves a PCKh detection rate of 92.7\% and outperformed other methods for single person pose estimation.
% \begin{table}[!ht]
% \begin{center}
% \begin{tabular}{|c|c|}
% \hline
%     Architecture&PCKh\\
% \hline\hline
%     \textbf{UniPose (ours)}&\textbf{92.7\%}\\
%     Deeply-Learned Models \cite{DeeplyLearned} & 92.3\%\\
%     Structure-Aware \cite{Multi-scale_structureAware} & 92.0\%\\
%     CPM \cite{CPM} & 88.5\%\\
% \hline
% \end{tabular}
% \caption{Pose estimation results using Percentage of Correct Keypoints to head (PCKh)
% for the MPII dataset \cite{MPII}}
% \end{center}
% \label{tab:MPII}
% \end{table}
Examples of pose estimation with UniPose in the MPII dataset are shown in Figure \ref{fig:MPII_sample}. 
These examples illustrate that UniPose deals effectively with occlusion, e.g. in the case of the horse rider.
% In the testing and pose detection for this dataset, the center position of the target individual is provided avoiding the decision of correct target for the pose estimation in case of multiple individuals in the frame.

\begin{table}[!ht]
\begin{center}
\begin{tabular}{|c|c|}
\hline
    \multirow{2}{*}{Method}&PCKh@0.5\\
    &for MPII\\
\hline\hline
    \textbf{UniPose (ours)}&\textbf{92.7\%}\\
    8-Stack HG \cite{zhang2019human} & 92.5\%\\
    Deeply-Learned Models \cite{DeeplyLearned} & 92.3\%\\
    Structure-Aware \cite{Multi-scale_structureAware} & 92.0\%\\
    Improvement Multi-Stage \cite{su2019improvement} & 90.1\%\\
    CPM \cite{CPM} & 88.5\%\\
\hline
\end{tabular}
\caption{Pose estimation results and comparisons with other methods 
% using Percentage of Correct Keypoints to head (PCKh)
for the MPII dataset.}
\end{center}
\label{tab:MPII}
\end{table}

\begin{figure*}[htbp]
\begin{center}
\includegraphics[width=1\linewidth]{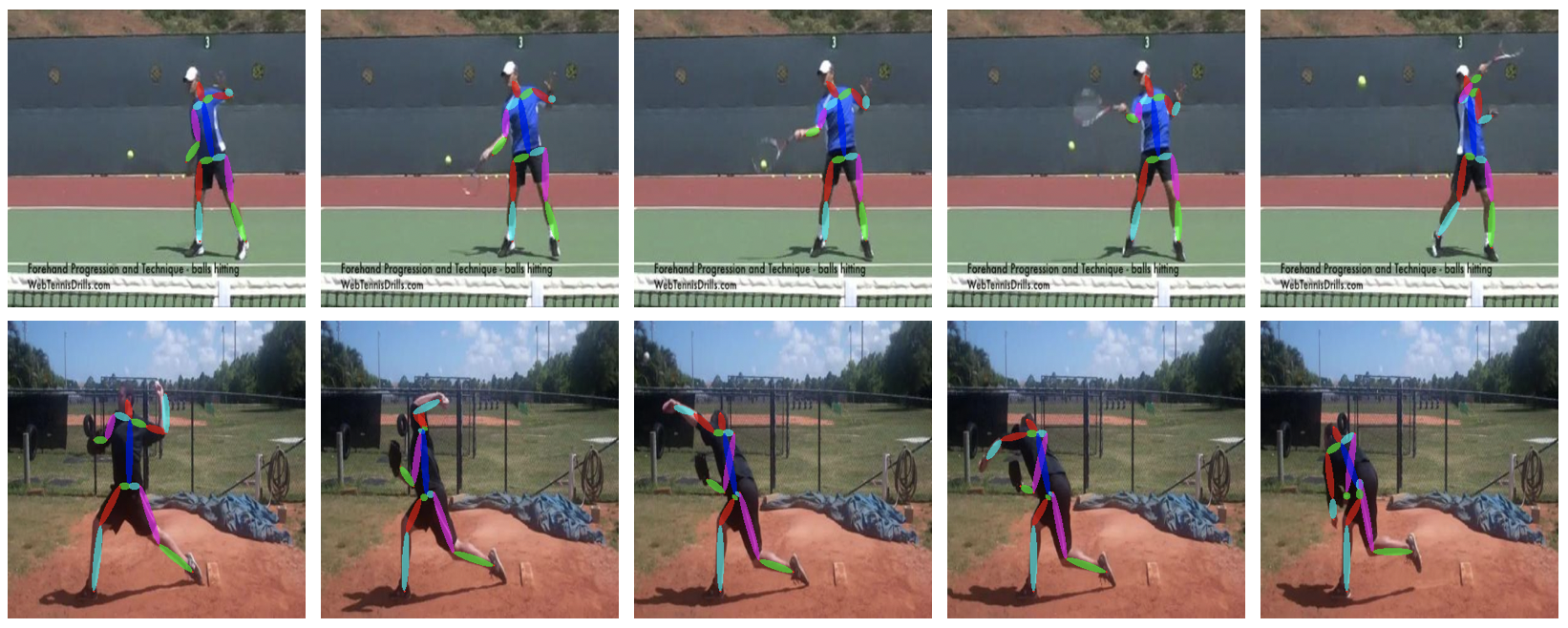}
\end{center}
  \caption{Pose estimation examples from the Penn Action dataset for a sequence of frames. }
\label{fig:PennAction_sample}
\end{figure*}

Table \ref{tab:PennAction} shows the results for our UniPose-LSTM in the Penn Action dataset \cite{PennAction}. Our results show a significant improvement over previous state-of-the-art methods by the application of UniPose-LSTM in the temporal mode with 5 consecutive frames. 
For this dataset, the results are reported as a correct detection when the predicted joint location lies within the provided bounding box, following the same procedure proposed by \cite{Articulated} and applied by \cite{LSTM-PM}.
Our method results in a 99.3\% detection rate, an improvement of 1.6\% over the next best result.

\begin{table}[!ht]
\begin{center}
\begin{tabular}{|c|c|}
\hline
    \multirow{2}{*}{Method}&PCK for\\
    &Penn Action\\
\hline\hline
    \textbf{UniPose-LSTM (ours)}&\textbf{99.3\%}\\
    LSTM-PM \cite{LSTM-PM} & 97.7\%\\
    CPM \cite{CPM} & 97.1\%\\
    Thin-Slicing \cite{Thin-Slicing} & 96.5\%\\
    N-best \cite{N-best} & 91.8\%\\
    Iqbal \cite{Iqbal} & 81.1\%\\
\hline
\end{tabular}
\end{center}
\caption{Pose estimation results and comparisons with other methods for the Penn Action dataset.}
\label{tab:PennAction}
\end{table}
Our UniPose network leverages  the memory capability of the LSTM by incorporating 5 consecutive frames. This feature enables a higher detection rate and consequently a more robust architecture against motion blur and occlusions in the image.

We experimented with different numbers of frames to evaluate the memory capability associated with the use of the LSTM. Table \ref{tab:LSTM} shows the accuracy gains observed from implementing LSTM for a number of frames ranging from 1 to 6. It is noticeable that the accuracy gains obtained by the LSTM plateaus as the number of frames reaches values of 5 or larger.

Examples of detections for the Penn Action dataset \cite{PennAction} are shown in Figure \ref{fig:PennAction_sample}. The examples selected are for situations of fast motion, showing every other frame in the sequence, so that significant differences are observed between the frames.

% \begin{table}[!ht]
% \begin{center}
% \begin{tabular}{|c|c|c|c|c|c|c|}
% \hline
%     LSTM&\multirow{2}{*}{1}&\multirow{2}{*}{2}&\multirow{2}{*}{3}&\multirow{2}{*}{4}&\multirow{2}{*}{5}&\multirow{2}{*}{6}\\
%     frames&&&&&&\\
% \hline\hline
%     PCK&98.4\%&98.6\%&98.8\%&99.1\%&99.3\%&99.3\%\\
% \hline
% \end{tabular}
% \end{center}
% \caption{UniPose-LSTM results based on PCK metric for different number of frames in the LSTM.}
% \label{tab:LSTM}
% \end{table}

\begin{table}[!ht]
\begin{center}
\begin{tabular}{|c|c|}
\hline
    Number of frames&PCK for\\
    in LSTM&Penn Action\\
\hline\hline
    1&98.4\%\\
    2&98.6\%\\
    3&98.8\%\\
    4&99.1\%\\
    5&99.3\%\\
    6&99.3\%\\
\hline
\end{tabular}
\end{center}
\caption{UniPose-LSTM results for the Penn Action dataset for different number of frames used by the LSTM.}
\label{tab:LSTM}
\end{table}

% \begin{figure}[htbp]
% \begin{center}
% \includegraphics[width=1\linewidth]{images/PennAction_sample.png}
% \end{center}
%   \caption{Pose estimation sample from the PennAction dataset for 5 frames.}
% \label{fig:PennAction_sample}
% \end{figure}

Table \ref{tab:BBC} shows results for the BBC Pose dataset, where pose is detected specifically for sign language. UniPose-LSTM significantly outperforms the older methods by achieving a PCKh of 98.9\%. In order to obtain results from another method for comparison, we trained CPM for the BBC Pose dataset, obtaining a PCK of 97.6\%, which is below the performance of UniPose-LSTM. 

\begin{table}[!ht]
\begin{center}
\begin{tabular}{|c|c|}
\hline
    \multirow{2}{*}{Method}&PCKh@0.5\\
    &for BBC Pose\\
\hline\hline
    \textbf{UniPose-LSTM (ours)}&\textbf{98.9\%}\\
    CPM \cite{CPM}&97.6\%\\
    Charles et al. \cite{AutomaticSign}&74.9\%\\
    Buehler et al. \cite{UpperBodySign}&67.5\%\\
\hline
\end{tabular}
\end{center}
\caption{Pose estimation results and comparisons with other methods for the BBC Pose dataset}
\label{tab:BBC}
\end{table}

\begin{figure}[htbp]
\begin{center}
\includegraphics[width=1\linewidth]{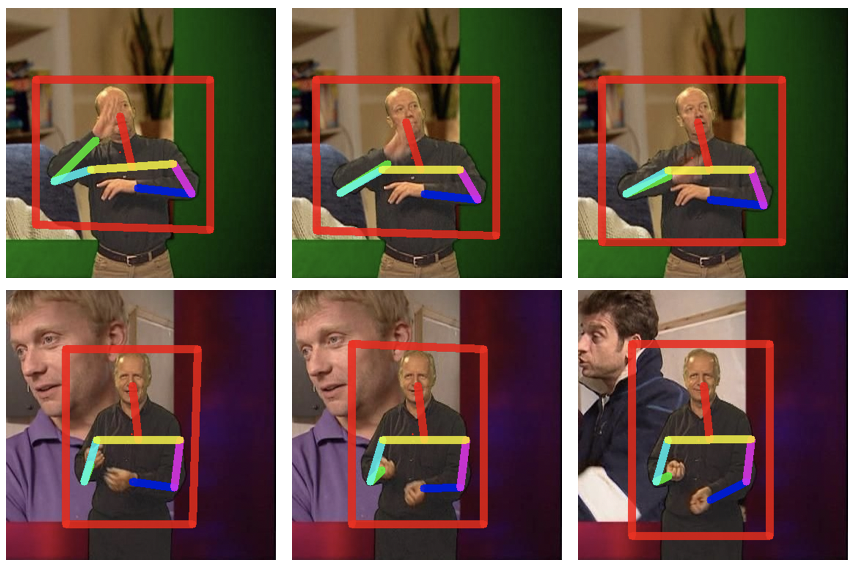}
\end{center}
  \caption{Pose estimation examples from BBC Pose dataset for a sequence of frames.}
\label{fig:BBC_detection}
\end{figure}

Figure \ref{fig:BBC_detection} shows examples of pose estimation and bounding box detections for subjects in the BBC dataset. Detections are shown for every other frame to illustrate different poses in the sequence.
Our network is able to efficiently detect the pose of the signers as well as generate the bounding box containing their signing area. 
% Figure \ref{fig:BBC_detection} shows the detections across 3 consecutive frames in the videos. 

% Similarly, when applying to the introduced PoseASL dataset for ASL, our UniPose method achieves a detection rate of 98.3\%, as shown in Table \ref{tab:NTID}.

% \begin{table}[!ht]
% \begin{center}
% \begin{tabular}{|c|c|}
% \hline
%     Architecture&PCKh\\
% \hline\hline
%     \textbf{UniPose (ours)}&\textbf{98.3\%}\\
%     CPM \cite{CPM}&69.6\%\\
% \hline
% \end{tabular}
% \end{center}
% \caption{PoseASL dataset pose results}
% \label{tab:NTID}
% \end{table}

% A sample for the pose estimation and bounding box for the subject from PoseASL dataset is shown in Figure \ref{fig:NTID_detection}.

% \begin{figure}[htbp]
% \begin{center}
% \includegraphics[width=1\linewidth]{images/NTID-detections.png}
% \end{center}
%   \caption{Pose estimation sample from PoseASL dataset}
% \label{fig:NTID_detection}
% \end{figure}

\section{Conclusion}
We presented the UniPose and UniPose-LSTM architectures for pose estimation in single images and videos, respectively. 
% Our framework incorporates our novel WASP module with waterfall architecture for multi-scale representations. 
% The video configuration of our UniPose with the incorporation of LSTM and a short addition of layers to the CNN results in state-of-the-art detections for video inputs.
The UniPose pipeline utilizes the WASP module that features a waterfall flow with a cascade of atrous convolutions and multi-scale representations. 
% This contributes to high performance evaluating a high number of
The large FOV of WASP obtains a better interpretation of the contextual information in the frame, and contributes to more accurately estimating the pose of the subject.
% Our methods achieve state of the art scores in the LSP, MPII, Penn Action and BBC Pose datasets using various metrics.

The results of UniPose and UniPose-LSTM demonstrated superior performance compared to state-of-the-art methods for several datasets, i.e., LSP, MPII, Penn Action and BBC Pose, using various metrics. 

Our framework shows promise for further use in a broader range of applications, including multiple person pose detection and 3D pose estimation.

{\small
\bibliographystyle{ieee_fullname}
\bibliography{egbib}
}

\end{document}